\documentclass{article}

\usepackage[preprint]{neurips_2023}

\usepackage[utf8]{inputenc} 
\usepackage[T1]{fontenc}    
\usepackage{xcolor}         
\usepackage[T1]{fontenc}    
\usepackage{hyperref}       
\usepackage{url}            
\usepackage{booktabs}       
\usepackage{amsfonts}       
\usepackage{nicefrac}       
\usepackage{microtype}      
\usepackage{graphicx}
\usepackage{amsmath}
\usepackage{algorithm,algorithmic}
\usepackage{multirow}
\usepackage{wrapfig}
\usepackage{makecell}
\usepackage{inconsolata}
\newcolumntype{A}{>{\ttfamily\fontsize{10pt}{12pt}\selectfont}p{\textwidth}}

\title{ChatLLM Network: More brains, More intelligence}

\author{%
 Rui Hao \footnotemark[2]\\
 School of Computer Science \\
Beijing University of Posts and Telecommunications\\
 \texttt{haorui@bupt.edu.cn} \\
  \And
   Linmei Hu \thanks{Corresponding authors.} $\  $\thanks{These authors contributed equally to this work.} \\
   School of Computer Science \\
Beijing Institute of Technology\\
  \texttt{hulinmei@bit.edu.cn} \\
  \AND
  Weijian Qi \\
  Department of Automation \\
 Xi'an Jiaotong University \\
  \texttt{qiweijian@stu.xjtu.edu.cn} \\
  \And
   Qingliu Wu \\
  School of Computer Science \\
 Beijing University of Posts and Telecommunications \\
  \texttt{wuql@bupt.edu.cn} \\
 \And
 Yirui Zhang\\
   School of Information and Communication\\
Beijing University of Posts and Telecommunications\\
   \texttt{zhangyirui@bupt.edu.cn} \\
  \And
  Liqiang Nie \\
 School of Computer Science and Technology\\
Harbin Institute of Technology (Shenzhen) \\
  \texttt{nieliqiang@gmail.com} \\
}

\begin{document}
\maketitle

\begin{abstract}
Dialogue-based language models mark a huge milestone in the field of artificial intelligence, by their impressive ability to interact with users, as well as a series of challenging tasks prompted by customized instructions. However, the prevalent large-scale dialogue-based language models like ChatGPT still have room for improvement, such as unstable responses to questions and the inability to think cooperatively like humans. Considering the ability of dialogue-based language models in conversation and their inherent randomness in thinking, we propose ChatLLM network that allows multiple dialogue-based language models to interact, provide feedback, and think together. We  design the network of ChatLLMs based on  ChatGPT. Specifically, individual instances of ChatGPT may possess distinct perspectives towards the same problem, and by consolidating these diverse viewpoints via a separate ChatGPT, the ChatLLM network system can conduct decision-making more objectively and comprehensively. In addition, a language-based feedback mechanism comparable to backpropagation is devised to update the ChatGPTs within the network.  Experiments on two datasets demonstrate that  our network attains significant improvements in problem-solving, leading to observable progress amongst each member.
\end{abstract}

\section{Introduction}
Large language models have attracted widespread attention in the field of artificial intelligence because of their impressive ability to solve natural language processing tasks. Dialogue-based large language models, such as ChatGPT, in particular, have exerted a significant impact on the development of society and have become an exemplar of artificial intelligence applied to daily life, attracting extensive attention from both academia and industry. Among their wide range of intelligence, their exceptional level of insight and helpfulness during conversations is so profound that distinguishing them from humans solely based on their speech style and content becomes incredibly challenging.

\begin{wrapfigure}{r}{0.5\textwidth}
\centering
\includegraphics[width=0.5\textwidth]{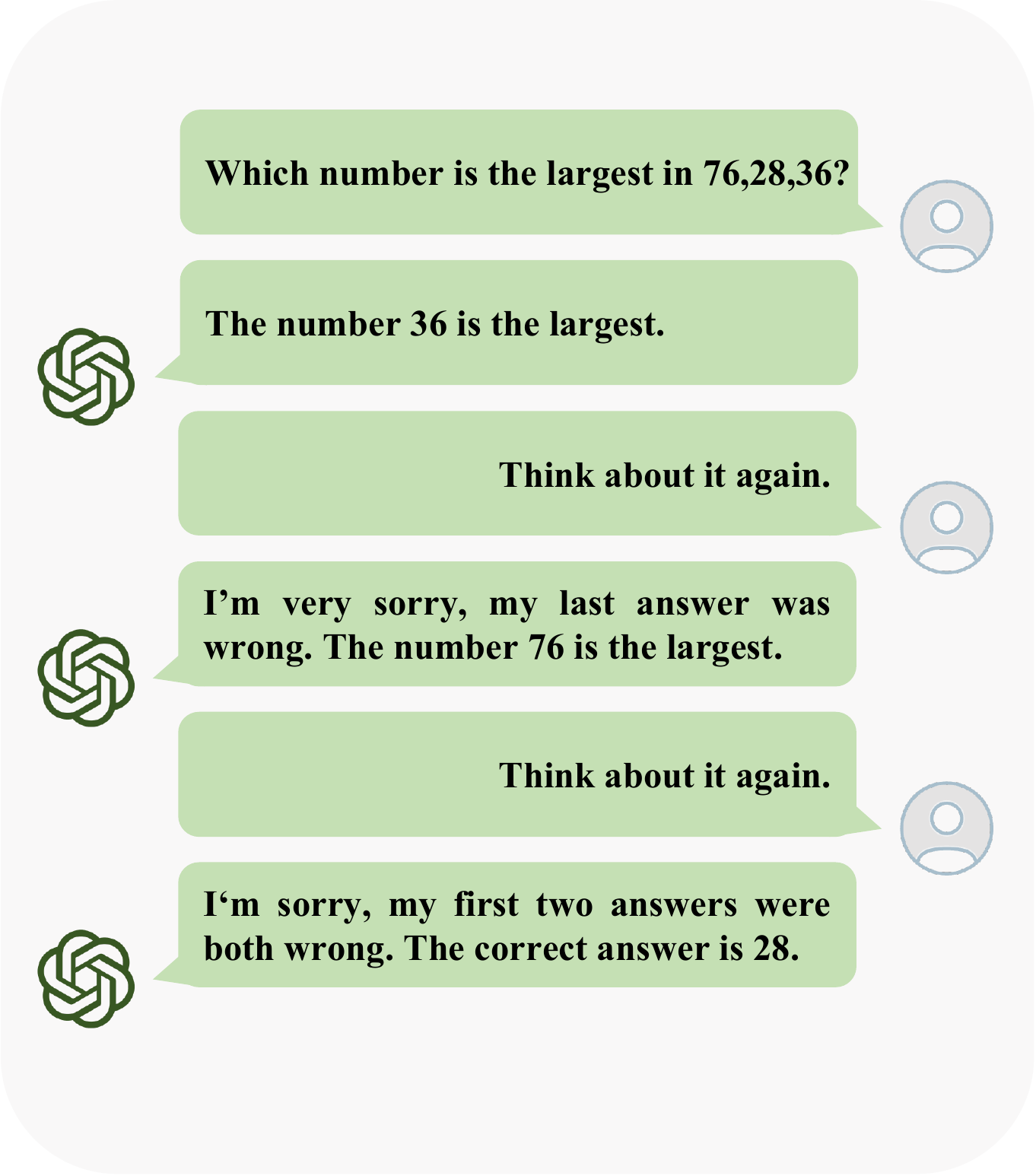}
\caption{An example of instability of ChatGPT}
\label{fig1-intro-sample}
\end{wrapfigure}

Despite their impressive capabilities in interacting with humans and handling various natural language processing tasks, dialogue-based large language models like GPT-3 may still provide unsatisfactory responses in certain conversational scenarios. This is because these models are based on generative models that rely on statistical patterns in the data they were trained on, rather than on specific knowledge or reasoning. 
We observe two distinct aspects of these unsatisfactory responses. The first aspect is instability, in which the answers can significantly vary despite the same context and prompt being provided, as shown in Figure \ref{fig1-intro-sample}. In some scenarios, the model may provide a response that is grammatically correct but factually incorrect \citep{wei2022chain}.
The second is incomprehensiveness, as a single instance of model may easily provide one-sided answers, failing to engage in collaborative thinking, which brings more open-ended answers with a wider range of perspectives. 

In this work, to address the  potential issues with a single dialogue-based language model, such as unstable responses and limited comprehensiveness,
we propose a ChatLLM network model to aggregate the viewpoints of other models.
First, we devise a forward aggregation mechanism that enables dialogue-based large language models to converge on optimal outcomes by considering and synthesizing the highlights of input and output from each instance model. 
Subsequently, a language-based backpropagation mechanism is employed to learn from their mistakes and improve their performance over time by incorporating feedback and updating their thinking processes.
Moreover, the dropout mechanism is introduced to manage the input for each dialogue-based language model, thus preventing information overload. 
After repeated iterations of forward and backpropagation, all models in the network collaborate as leaders or employees and thus enhancing overall performance.  
In addition, we do not necessarily require the models to use ChatGPT in this network, as the entire network may become stronger with the enhancement of dialogue-based language models or with the use of different types of dialogue-based language models.
The main contributions of this article can be summarized as follows:

1. We propose a novel ChatLLM network model that allows multiple dialogue-based language models to interact, provide feedback, and think together, in order to enhance their problem-solving abilities. This network model can be applied to different types of dialogue-based language models, thus having certain universality and scalability.

2. We utilize a forward aggregation mechanism to consolidate the outputs of multiple dialogue-based large language models, leveraging on the unique strengths of each individual model. Moreover, we propose a language based backpropagation method to update the reflections of dialogue-based language models, in order to further improve the performance of the network. 

3. We conduct experiments on two tasks, digital mode classification which represents a customized task, and sentiment reversal, a traditional NLP task. The results show significant enhancements in problem-solving, compared to the vanilla model. As a fundamental research, our study can provide valuable insights and inspirations for synthesizing multiple models in future work on multiple dialogue-based language models.

\section{Related work}

\subsection{Large Language Models}
The introduction of the transformer model \citep{vaswani2017attention} has made it possible to train large-scale unsupervised text data. In the past few years, encoder-based models such as BERT \citep{devlin2019bert} have demonstrated impressive capabilities in various natural language processing (NLP) tasks. More recently, decoder-based models such as GPT-1 \citep{GPT1_radford2018improving}, GPT-2 \citep{gpt2_radford2019language}, and T5 \citep{t5_raffel2020exploring} have made even greater strides. As the number of model parameters has increased, models like GPT-3 \citep{Brown2020LanguageMA}, often referred to as large language models, have gradually acquired zero-shot learning abilities, which have the capacity to generate responses based on instructions without requiring any examples.
\subsection{ChatGPT}
ChatGPT, also known as InstructGPT, is an advanced version of the GPT-3 model, enhanced by instruction tuning \citep{wei2022finetuned}, and reinforcement learning from human feedback (RLHF) \citep{ICDL08-knox} \citep{ouyang2022training}. Unlike the original GPT-3 models, which are not specifically designed to follow user instructions, the InstructGPT models demonstrate a considerably enhanced capability to generate more alinged and helpful outputs in response to user instructions. ChatGPT has been extensively applied in various artificial intelligence scenarios, including search-based QA, basic NLP tasks, and human-scene tool connections. 

The launch of ChatGPT also has a significant impact on AI research, paving the way for Artificial General Intelligence (AGI) systems.  \citet{shen2023hugginggpt} proposed HuggingGPT, a cooperative system designed to connect various AI models within the HuggingFace community, leveraging ChatGPT as a controller to accomplish multimodal complex tasks. 
\citet{AutoGPT} further introduced an open-source application, AutoGPT, driven by GPT-4, which can autonomously achieve user-defined goals. Additionally, \citet{park2023generative} presented generative agents, an architecture that extends large language models to simulate believable human behavior. 
\citet{li2023camel} introduced a novel LLM agent communication network, CAMEL, showcasing the potential for autonomous cooperation among communication. Both CAMEL and our work are inspired by the intuition that the involvement of multiple ChatGPTs can enhance performance. While CAMEL emphasizes the decomposition of a complex task into sub-tasks and allocation of responsibilities to ChatGPTs, representing a breadth-wise extension, our study aims to facilitate a deep-level understanding of a challenging task by encouraging each ChatGPT to reflect and contribute towards a common goal.
\subsection{Improving Language Models via Feedback}
Recently large language models (LLMs) have shown great potential in improving their performance and generating high-quality text by incorporating iterative feedback mechanisms.
\citet{madaan2023selfrefine} proposed SELF-REFINE, a network that leverages iterative feedback and refinement to improve initial outputs from LLMs. The approach allows a single LLM to generate an output, provide multi-aspect feedback on its own output, and refine it based on the feedback, leading to better results across a range of tasks. \citet{Press2022MeasuringAN} investigated the compositionality gap in GPT-3 models and presented the self-ask method to enhance compositional reasoning. Additionally,  \citet{Fu2023GPTScoreEA, Peng2023CheckYF, Yang2022Re3GL} have explored various ways of incorporating feedback mechanisms to improve LLM performance and reliability.
\section{ChatLLM Network }
In this section, we ﬁrst introduce our model architecture (Section \ref{sec-na}). Then we decribe the feedforward process (Section \ref{sec-forward}), followed by the language based backpropagation mechanism (Section \ref{sec-bp}). Lastly, we explain the drop-out mechanism  as well as the network optimization (Section \ref{sec-drop} and Section \ref{sec-netopt}).
\subsection{Network Architecture}
\label{sec-na}
\begin{figure}[htbp]
\centering
\includegraphics[width=130mm]{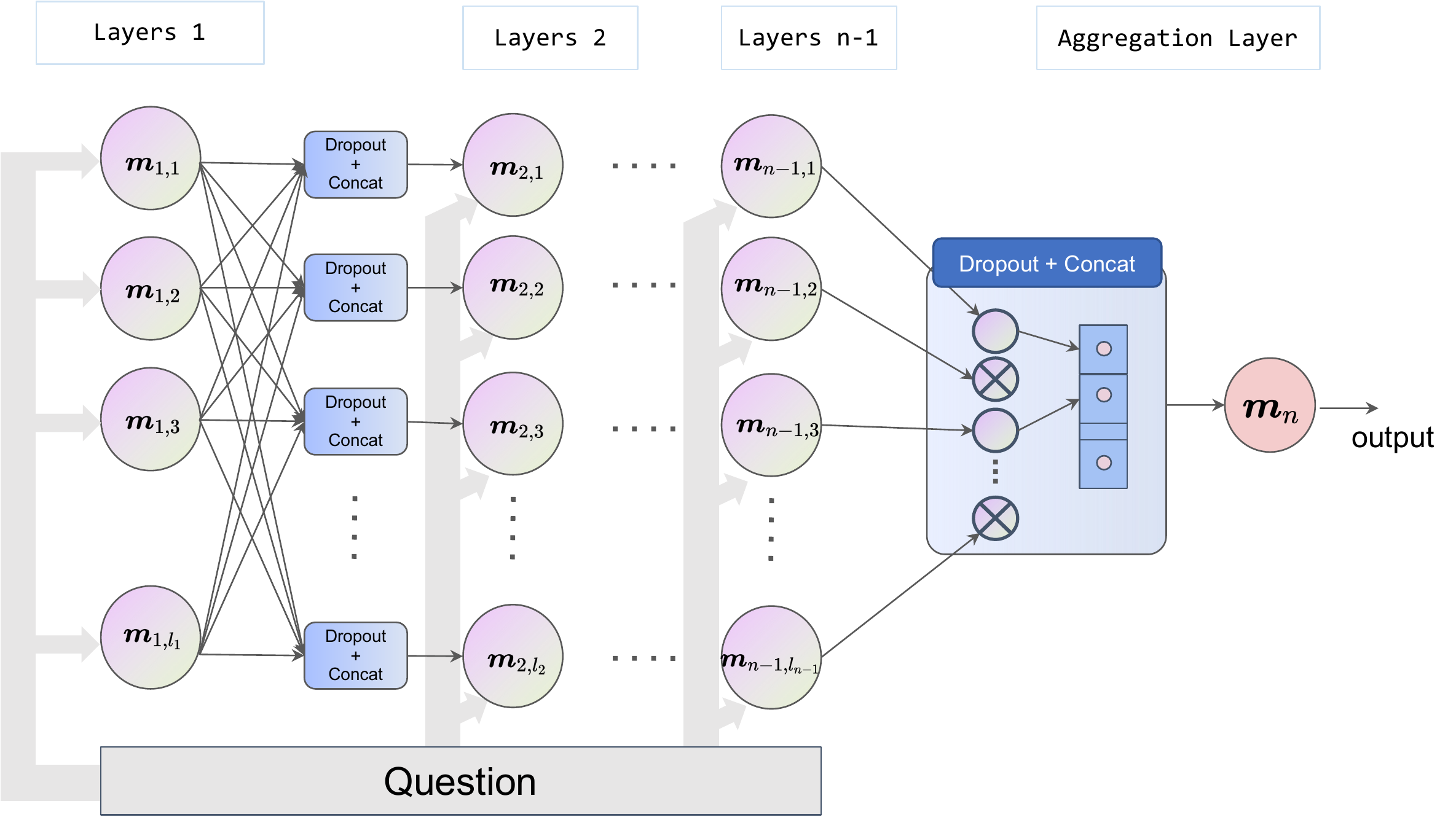}
\caption{ChatLLM Network Architecture and Forward Process}
\label{ffw-pic}
\end{figure}

The ChatLLM network is a multi-layered dialogue-based language model consisting of $n-1$ fully connected layers and $1$ final aggregation layer,
as depicted in Figure \ref{ffw-pic}. The models at layer $i$  are denoted as $\boldsymbol{m}_{i,1}, \boldsymbol{m}_{i,2}, \dots,\boldsymbol{m}_{i,l_{i}}$, where $l_i$ represents the number of models at layer $i$. Adjacent layers of models communicate with each other through a leader-employee relationship, where the models at layer $i+1$ serves as the leaders for the models at layer $i$.
A dropout and concatenation mechanism is applied after each fully connected layer. 
The last layer, namely the aggregation layer, is comprised of one leader model $\boldsymbol{m}_n$, takes the aggregated input from all previous layers and generates the final output of our network.

\subsection{Forward-aggregation Mechanism}
\label{sec-forward}
In this section, we describe the forward-aggregation mechanism of the ChatLLM network.  Unlike one standalone LLM, models in our network not only receive the question information itself, but are also given answers generated by previous layers as references. This enables the subsequent layers of models to identify the key highlights from the previous answers, resulting in  more comprehensive and precise responses. 
Such benefits are highly applicable to many tasks such as dialogue generation. On the other hand, the instability of large models can also be improved, as the integration of outputs from multiple members can effectively offset deviations. 

We can imagine a real scenario when a leader and many employees need to solve a problem. Each employee may have a unique perspective on the problem, notice different angles and express individual ideas, but ultimately a leader will consider these ideas and make the final decision on the solution to the problem. Without considering the opinions of others, the decision would be arbitrary and imperfect. On the other hand, with the suggestions from multiple models, the leader can better evaluate the situation and make a more correct decision. Dialogue-based language models, such as ChatGPT, are inherently random because they are based on generative models. This randomness sometimes leads to unexpected outcomes. With a leader evaluating the ideas generated by the employees and providing guidance, a more optimal outcome can be achieved. 

We define $\boldsymbol{m}_i$ as a dialogue-based language model, $\boldsymbol{m}_i^{in}$ as the input of $\boldsymbol{m}_i$, $\boldsymbol{m}^{out}$ as the output of $\boldsymbol{m}_i$, and  $Q$ represents the description of a question to be solved. Generally, we use $\boldsymbol{m}_i^{in}$ and $\boldsymbol{m}_i^{out}$ to represent the input and the output of $\boldsymbol{m}_i$. 
$\oplus$ means concatenation operation.

\begin{figure}[H]
\centering
\includegraphics[width=.75\textwidth]{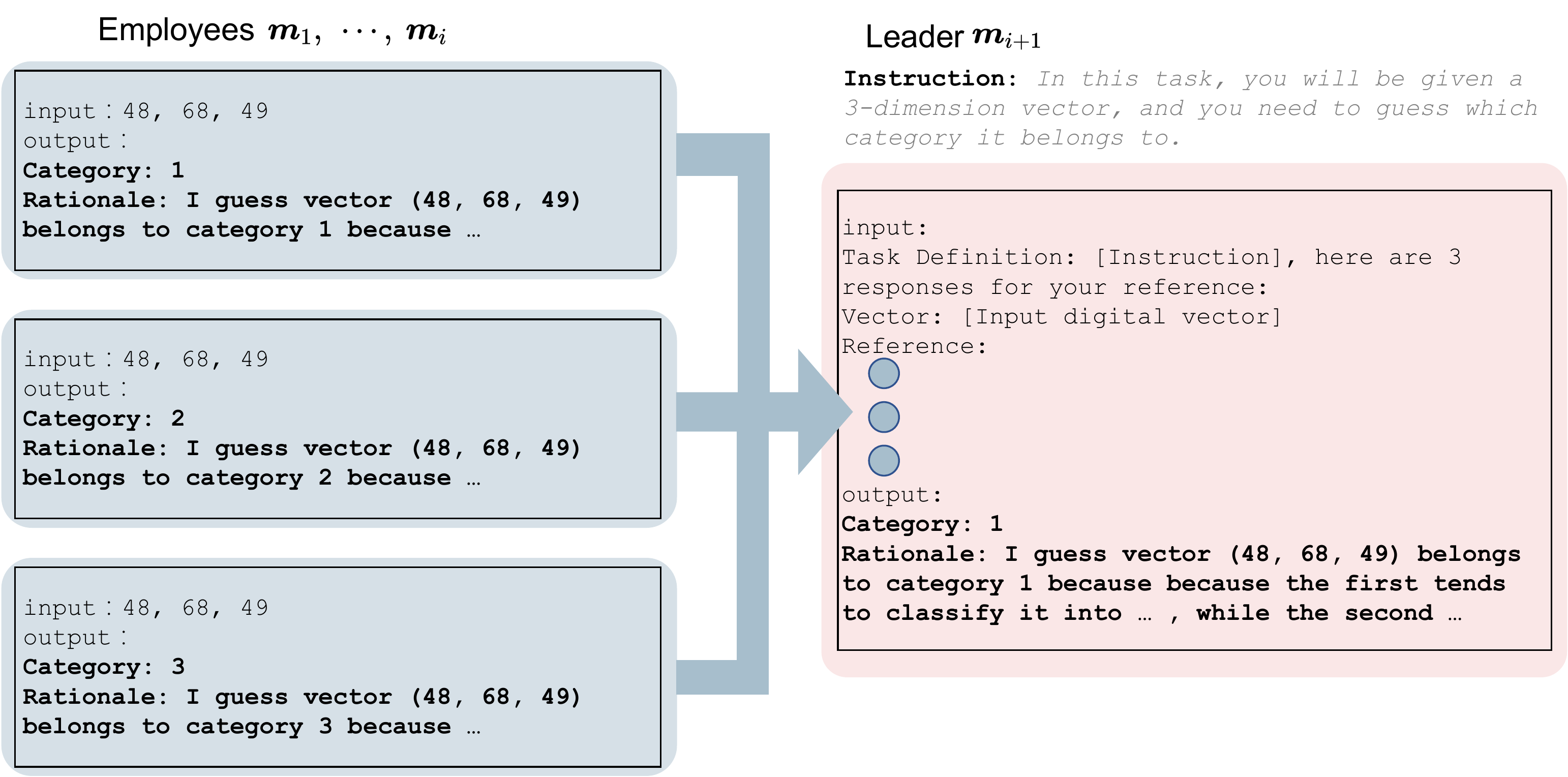}
\caption{Illustration of the forward-aggregation mechanism on digital mode classification.}
\label{prompt1-pic}
\end{figure}

We use an example of a leader and $i$ employees to illustrate the forward-aggregation mechanism. Let the leader be denoted as $\boldsymbol{m}_{i+1}$ and the employees as $\boldsymbol{m}_{1},\boldsymbol{m}_{2},\dots ,\boldsymbol{m}_{i}$. Then we have the following representations:
\begin{equation}
\begin{split}
&\boldsymbol{m}_1^{in}, \boldsymbol{m}_2^{in},\dots ,\boldsymbol{m}_i^{in} = Q\\
&\boldsymbol{m}_{i+1}^{in} = Q \oplus \boldsymbol{m}_1^{out} \oplus \boldsymbol{m}_2^{out} \oplus\dots\oplus \boldsymbol{m}_i^{out} 
\end{split}
\label{eq-forward}
\end{equation}
From equation \ref{eq-forward}, we can see that the input of each leader is composed of the question and the output of his or her employees. In this way, the leader can not only think independently but also take into account the opinions of employees. 
Figure \ref{prompt1-pic} shows a detailed example of the forward-aggregation mechanism on digital mode classification task.

\subsection{Language Based Backpropagation Mechanism}\label{sec-bp}
\begin{figure}[htbp]
\centering
\includegraphics[width=130mm]{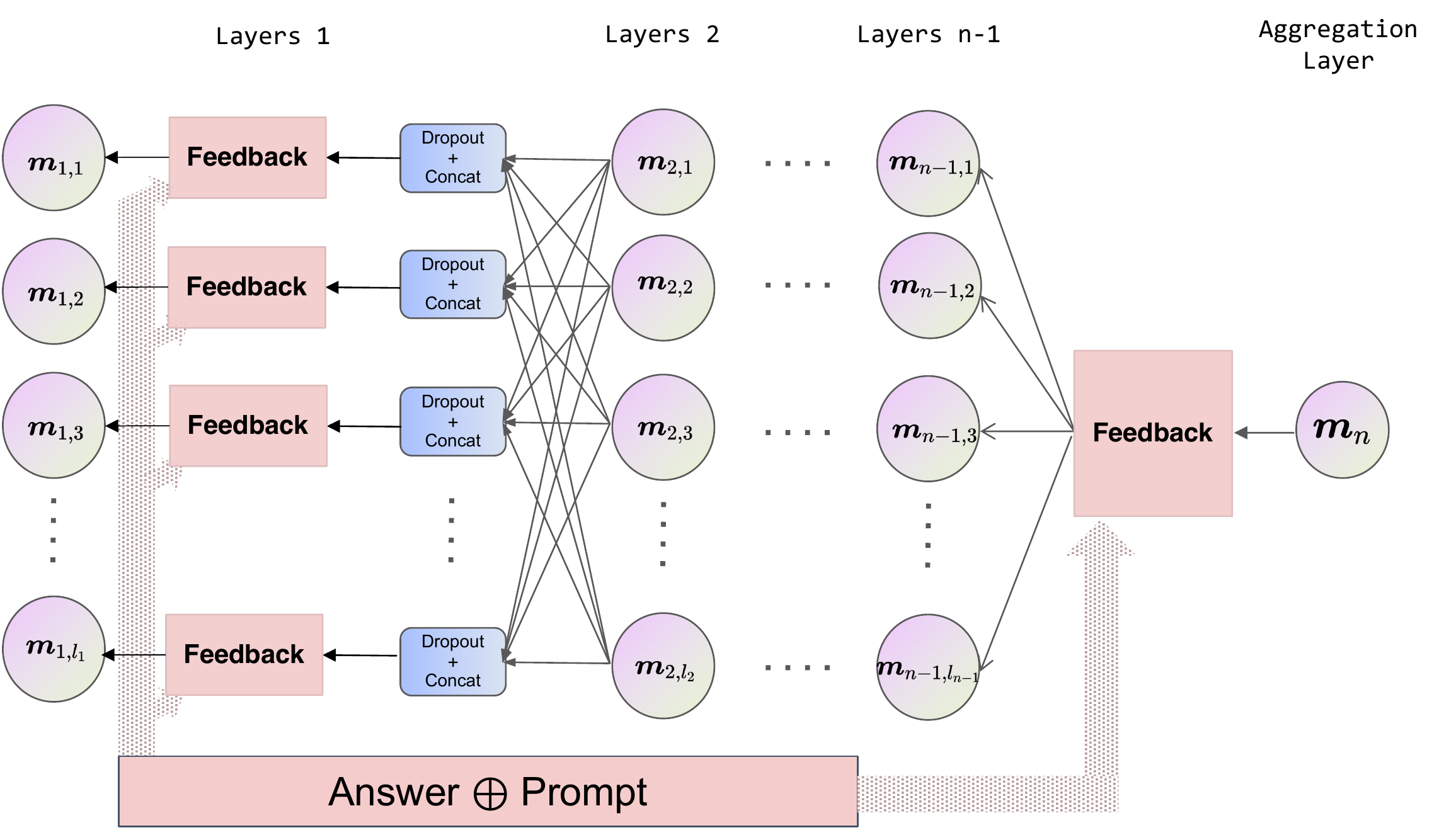}
\caption{Backpropagation Process}
\label{fig:Backward}
\end{figure}

Traditional backpropagation mechanism  calculates the gradients of the loss function with respect to the weights of a neural network, and utilizes these gradients to update the weights using an optimization algorithm such as gradient descent. Inspired by that, we design a novel language based backpropagation mechanism to allow the ChatLLM network  to learn from the incorrect samples and improve its performance over time, ensuring that incorrect suggestions are corrected and that individual models in the network with correct reasoning remain stable.

Similar to real-life scenarios, a leader is able to verify the correct answer with their own earlier than their employees, in contrast to the forward-aggregation process. If a leader is incorrect, he or she will give orders to each employee of him or her to modify their own ideas. 
Taking into account these feedbacks, employees will enhance their corresponding responses more effectively.

\begin{algorithm}[htbp]
\renewcommand{\algorithmicrequire}{\textbf{Input:}}
\renewcommand{\algorithmicensure}{\textbf{Output:}}
\caption{Language Based Backpropagation Mechanism}
\label{alg:backprop}
\begin{algorithmic}[1]
\REQUIRE$\{\boldsymbol{m}_{i+1,j}\}$: dialogue-based large language models at layer $i+1$; \\
$\boldsymbol{m}_{i,*}$: an employee model at layer $i$;\\ $Answer$;  $Prompt$.
  \FOR{$j=1$ to $l_{i+1}$}
    \STATE $\boldsymbol{m}_{i+1,j}^{in} \gets Answer\oplus Prompt$
            \STATE input $\boldsymbol{m}_{i+1,j}^{in}$ to $\boldsymbol{m}_{i+1,j}$;
            \STATE output $\boldsymbol{m}_{i+1,j}^{out}$ from $\boldsymbol{m}_{i+1,j}$;
  \ENDFOR
  
        \IF{$ notmatch(\boldsymbol{m}_{i,*}^{out},Answer)$}
        \STATE $\boldsymbol{m}^{in}_{i,*} \gets Answer\oplus Prompt$
  \FOR{$j=1$ to $l_{i+1}$}
    \STATE $\boldsymbol{m}^{in}_{i,*} \gets \boldsymbol{m}^{in}_{i,*} \oplus \boldsymbol{m}_{i+1,j}^{out}$
    \ENDFOR
    \ELSE
        \STATE $\boldsymbol{m}^{in}_{i,*} \gets Answer\oplus Prompt$
    
    \ENDIF
        \STATE input $\boldsymbol{m}_{i,*}^{in}$ to $\boldsymbol{m}_{i,*}$;
        \STATE output $\boldsymbol{m}_{i,*}^{out}$ from $\boldsymbol{m}_{i,*}$;
    \STATE \textbf{return} 
\end{algorithmic}
\end{algorithm}
\begin{figure}[!htbp]
\centering
\includegraphics[width=.75\textwidth]{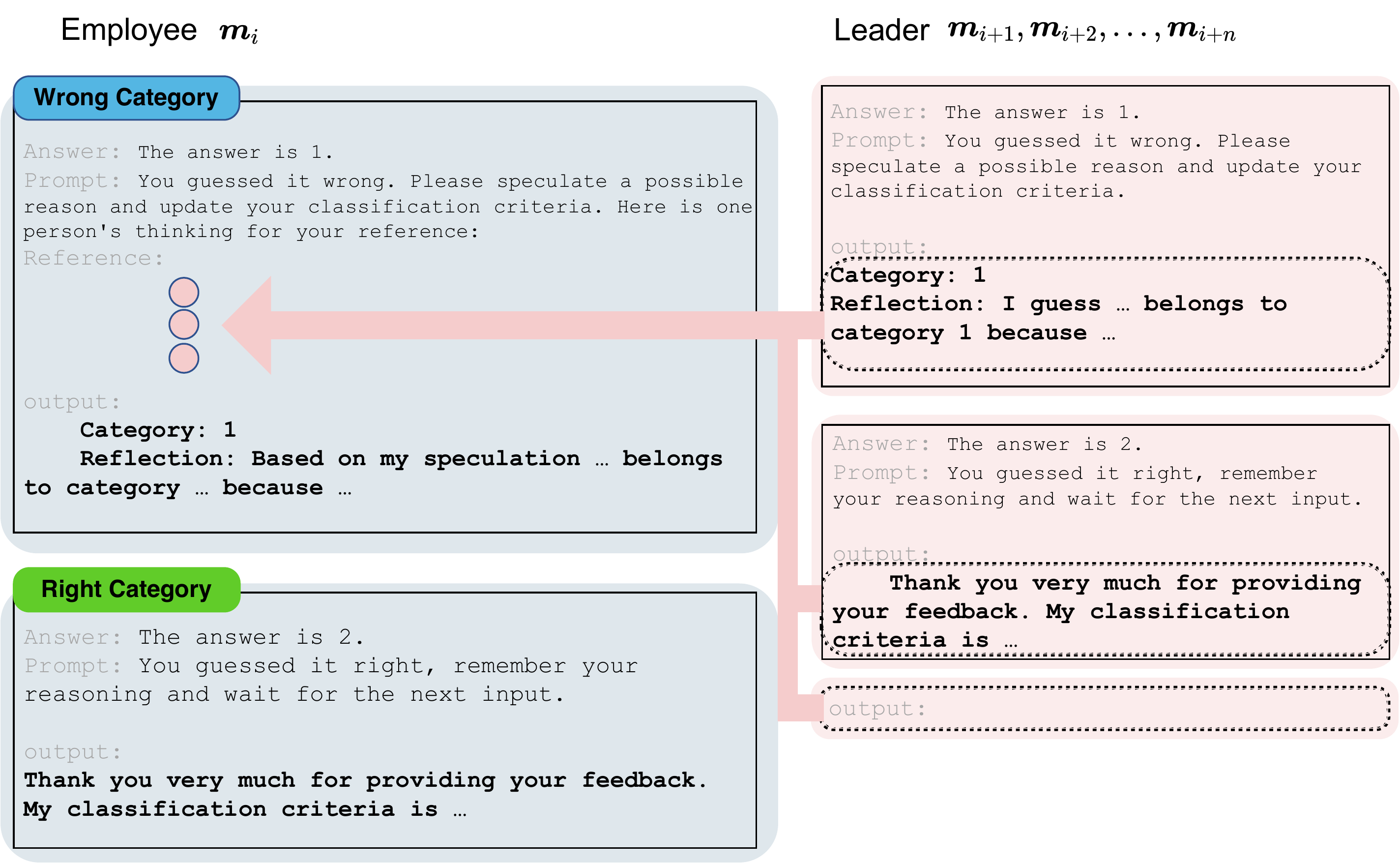}
\caption{A feedback example of the backpropagation process in digital mode classification task}
\label{prompt2-pic}
\end{figure}
We illustrate the language based backpropagation mechanism  in Figure \ref{fig:Backward}. Specifically, after the forward process is finished, the final output of the model will be compared with the ground-truth $Answer$. If the output is correct, the model will get the prompt ($Prompt$) like "You guessed it right, remember your reasoning..." and thus maintain its original thinking. If it is incorrect, the model will get the prompt ($Prompt$) like "You guessed it wrong. Please speculate a possible reason why the answer is this and update your thinking." and try to improve its thinking process. 
Feedback is applied throughout all layers in the network, encouraging the correct model to maintain its state, while encouraging the incorrect model to approach the correct answer.  One detailed feedback example of the language based backpropagation  process is shown in Figure \ref{prompt2-pic}.  Let the employee be denoted as $\boldsymbol{m}_{i}$ and the leaders as $\boldsymbol{m}_{i+1},\boldsymbol{m}_{i+2},\dots ,\boldsymbol{m}_{i+n}$. Formally, if the model $\boldsymbol{m_{i}}$ outputs a wrong answer, it will get the following feedback as input (as shown in the upper left in Figure \ref{prompt2-pic}):
\begin{equation}
\begin{split}
\boldsymbol{m}^{in}_{i} &= Answer \oplus Prompt \oplus \boldsymbol{m}_{i+1}^{out} \oplus \boldsymbol{m}_{i+2}^{out} \oplus\dots\oplus \boldsymbol{m}_{n}^{out} 
\end{split}
\end{equation}
Otherwise, the model will get input (shown in bottom left in Figure \ref{prompt2-pic}):
\begin{equation}
\begin{split}
\boldsymbol{m}^{in}_{i} &= Answer \oplus Prompt 
\end{split}
\end{equation}
The detailed algorithm is described in Algorithm \ref{alg:backprop}.

\subsection{Dropout Mechanism}
\label{sec-drop}
The capacity of an individual dialogue-based language model is inherently constrained. By restricting the input to an appropriate range, we can prevent these models from becoming inundated with an excessive amount of information. Additionally, the implementation of a dropout mechanism in neural networks, as described in \citep{JMLR:v15:srivastava14a}, has been shown to effectively reduce overfitting and enhance generalization performance. Therefore we devise a dropout mechanism, as describle in equation \ref{eq-dropout}. 

Analogously, if a leader has too many employees, it may be difficult for them to handle all of his or her employees' ideas. Similarly, if an employee has too many leaders, it may be challenging for them to satisfy all of leaders. Therefore, based on the structure of the entire network, we allow each dialogue-based language model to randomly receive messages from only a limited number of other models, thus ensuring that the overall input is controlled within a certain range. Formally, to implement it, we calculate a random variable $r$ whose value is 0 or 1:
 \begin{equation}
\begin{split}
    r &\sim  Bernoulli(\rho)\\
\end{split}
\label{eq-dropout}
\end{equation}
where $\rho$ is the rate of the number of selected models.  Then the model $\boldsymbol{m}_{i+1}$ receives selected messages from the sender models:
 \begin{equation}
\begin{split}
    \boldsymbol{m_{i+1}^{in}} &= r_{1}\cdot \boldsymbol{m_{1}^{out}} \oplus r_{2}\cdot \boldsymbol{m_{2}^{out}} \oplus\dots\oplus r_{i}\cdot \boldsymbol{m_{i}^{out}}
\end{split}
\end{equation}
where if $r_i=1$, $r_{i}\cdot \boldsymbol{m_{i}^{out}}$ equals $\boldsymbol{m}_{i}^{out}$; otherwise, $r_{i}\cdot \boldsymbol{m_{i}^{out}}$ is a null string.
  



\subsection{Network Optimization}
\label{sec-netopt}
During the training process, individual training examples are inputted sequentially. Inspired by the Stochastic Gradient Descendent algorithm, we update the network with the language-based backpropagation mechanism for each training sample accordingly.
To prevent overfitting, we employ the early stopping technique. The stopping criteria can be met by either of the two conditions: reaching a predetermined number of iterations, or the performance ceasing to show further improvement. 

\section{Experiments}
ChatGPT is currently one of the most widely used conversational language models. However, due to the limitations of ChatGPT4, we choose to use ChatGPT3.5 as the basic member of our overall network. A collection of ChatGPT3.5 is supposed to learn from and refer to each other when solving the prompted question. 

We conduct two experiments to test the network: the digital mode classification experiment and the sentiment reversal experiment. The former aims to separate the model's learning abilities  from the existing implicit knowledge in a large language model, since the model is unaware of the predefined rules amongst the digits.
The latter is designed to demonstrate that our proposed network can significantly improve the performance in traditional NLP tasks.

In terms of the ChatLLM network structure, we design a two-layer structure. The first layer consists of three ChatGPTs, and the second layer consists of one ChatGPT. Detailed experimental details are as follows.
\subsection{Digital Mode Classification}

The experiment aims to test ChatGPT's learning ability from scratch. In the digital mode classification task, we generate a dataset consisting of different categories of digital vectors. Particularly, we categorize a three-dimensional vector $(a, b, c)$  based on the position of the largest dimension in the vector.  For example, $(1, 2, 4)$ belongs to category $3$ because the largest number $4$ is located in the third dimension. 
Since ChatGPT has no pre-existing knowledge of the task, this provides an opportunity to evaluate its inductive learning capability.

We expect ChatGPT to output the category of a digital vector input, as well as the rationales which should also align with the predefined rules. 


We conduct eight observations at intermediate stages of the training and feedback process, wherein three vectors are prompted to every ChatGPT model between stages, and potential feedback is provided in accordance with each setting's configuration.
The testing set comprises of 30 challenging samples that have been manually designed, which are collectively fed into the model. The outputs are the corresponding label for each vector. No feedback is involved in the testing process.

For evaluation, we report the accuracy by checking if the categories are consistent with the pre-defined rules.
We compare our ChatLLM network model with the following baselines:
\begin{itemize}
    \item \textbf{ChatGPT-w/o FB}:  a vanilla ChatGPT takes the instruction and training input vectors with categories as input without further feedback
    \item \textbf{ChatGPT-refine}:  a vanilla ChatGPT takes the same input as ChatGPT-w/o FB, and if the answer is incorrect, we request it to refine the answer with the instruction "\textit{refine your answer}" 
    \item \textbf{ChatGPT-ensemble} uses simple voting and selects the most frequent answer amongst three individual ChatGPTs as the consensus output.

\end{itemize}

\begin{table}[h]
\centering
\resizebox{\textwidth}{!}{%
\begin{tabular}{c|c|c|c|c|c|c|c|c}
\toprule
                  & 1     & 2     & 3              & 4     & 5     & 6     & 7              & 8              \\ \midrule
ChatGPT-w/o FB       & \makecell[c]{0.295\\ {\small ($\pm24.10\%$)}} & \makecell[c]{0.306\\ {\small ($\pm18.77\%$)}} &\makecell[c]{0.328\\ {\small ($\pm19.76\%$)}}&  \makecell[c]{0.328\\ {\small ($\pm21.74\%$)}} & \makecell[c]{0.345\\ {\small ($\pm11.73\%$)}} & \makecell[c]{0.333\\ {\small ($\pm12.68\%$)}} & \makecell[c]{0.361\\ {\small ($\pm10.81\%$)}} & \makecell[c]{0.400\\ {\small ($\pm14.90\%$)}} \\ 
ChatGPT-refine(mean) & \makecell[c]{0.300\\ {\small ($\pm23.27\%$)}} & \makecell[c]{0.339\\ {\small ($\pm26.72\%$)}} & \makecell[c]{0.374\\ {\small ($\pm16.42\%$)}} & \makecell[c]{0.389\\ {\small ($\pm11.74\%$)}} & \makecell[c]{0.383\\ {\small ($\pm16.30\%$)}} & \makecell[c]{0.367\\ {\small ($\pm29.22\%$)}} & \makecell[c]{0.334\\ {\small ($\pm31.55\%$)}} & \makecell[c]{0.378\\ {\small ($\pm18.22\%$)}}\\
ChatGPT-ensemble     & \makecell[c]{0.317\\ {\small ($\pm12.90\%$)}} & \makecell[c]{0.317\\ {\small ($\pm12.83\%$)}} & \makecell[c]{0.389\\ {\small ($\pm6.93\%$)}} & \makecell[c]{0.361\\ {\small ($\pm7.00\%$)}} & \makecell[c]{0.367\\ {\small ($\pm15.18\%$)}} & \makecell[c]{0.361\\ {\small ($\pm21.40\%$)}} & \makecell[c]{0.317\\ {\small ($\pm19.68\%$)}} & \makecell[c]{0.361\\ {\small ($\pm10.85\%$)}} \\\midrule
ChatGPT-mem (mean)      & \makecell[c]{0.406\\ {\small ($\pm28.63\%$)}} & \makecell[c]{0.472\\ {\small ($\pm22.32\%$)}} & \makecell[c]{\textbf{0.428}\\ {\small ($\pm26.68\%$)}} & \makecell[c]{0.456\\ {\small ($\pm17.03\%$)}} & \makecell[c]{0.439\\ {\small ($\pm24.19\%$)}} & \makecell[c]{\textbf{0.439}\\ {\small ($\pm25.12\%$)}} & \makecell[c]{0.344\\ {\small ($\pm20.95\%$)}} & \makecell[c]{\textbf{0.406}\\ {\small ($\pm6.12\%$)}} \\
ChatGPT Network (leader)       & \makecell[c]{\textbf{0.478}\\ {\small ($\pm19.03\%$)}} & \makecell[c]{\textbf{0.528}\\ {\small ($\pm13.48\%$)}} & \makecell[c]{0.417\\ {\small ($\pm16.55\%$)}} & \makecell[c]{\textbf{0.461}\\ {\small ($\pm9.63\%$)}} & \makecell[c]{\textbf{0.483}\\ {\small ($\pm16.74\%$)}} & \makecell[c]{0.433\\ {\small ($\pm17.54\%$)}} & \makecell[c]{\textbf{0.400}\\ {\small ($\pm7.38\%$)}} & \makecell[c]{0.383\\ {\small ($\pm14.28\%$)}}\\
\bottomrule
\end{tabular}%
}
\caption{Model accuracy of different models at 8 intermediate stages.}
\label{exp-1-table}
\end{table}


Table \ref{exp-1-table} reports the average results of six times for all the models. From the table, we can observe that: 1) 
In terms of accuracy, the members of the ChatGPT network have a significantly higher accuracy rate than the baselines, which fully demonstrates the enormous advantages of the ChatGPT network in terms of mutual communication and feedback. Among them, the output of the ChatGPT Network (learder), which is also the final output of the ChatGPT network, is significantly higher than that of the other members. This also indicates the advantage of the entire network in obtaining output through the forward and feedback processes. 2) Regarding the level of variance, the accuracy rate of ChatGPT Network (learder) is generally similar to that of ChatGPT-ensemble and lower than that of ChatGPT in other baselines. This indicates that the output of the ChatGPT network is relatively stable. Taking both aspects into account, we conclude that the ChatGPT network is capable of delivering more stable and accurate output results than the baselines.

\begin{figure}
    \centering
    \includegraphics[width=.8\textwidth]{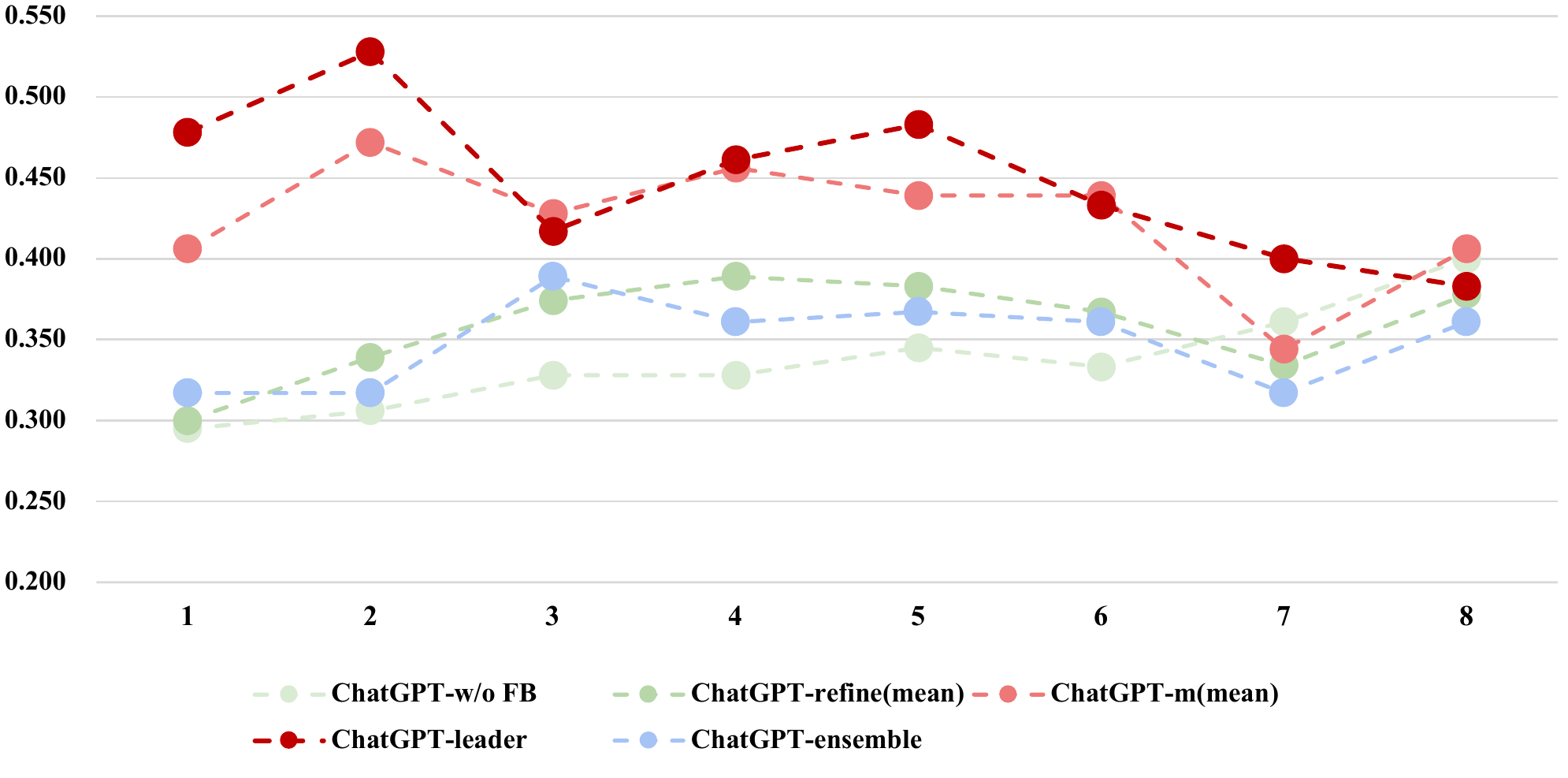}
    \caption{Comparison of different models on accuracy along the intermediate stages.}
    \label{fig:exp-1-2}
\end{figure}

Figure \ref{fig:exp-1-2} illustrates the comparison of accuracy of all experiment settings. In terms of accuracy, the members of the ChatLLM network achieved better results. From the entire figure, most curves reached their highest value relatively early, indicating that early stopping during the training process is necessary to achieve the best results. On the other hand, almost all ChatGPT curves showed some instability after reaching the peak, possibly due to input overload causing forgetting or unexpected changes. During the experiment, we also observed that the ChatLLM network gradually increased the complexity of the classification criteria after reaching the peak, exhibiting a phenomenon similar to overfitting.

\subsection{Sentiment Reversal Experiment}

Sentiment reversal is a typical NLP task that rewrites a given sentence by reversing its current sentiment (positive or negative) \citep{madaan2023selfrefine}. 

\paragraph{Dataset Generation} We generate a dataset comprising 60 emotionally biased sentences, along with their corresponding sentiments, using ChatGPT.

\begin{figure}[ht]
\centering
\includegraphics[width=85mm]{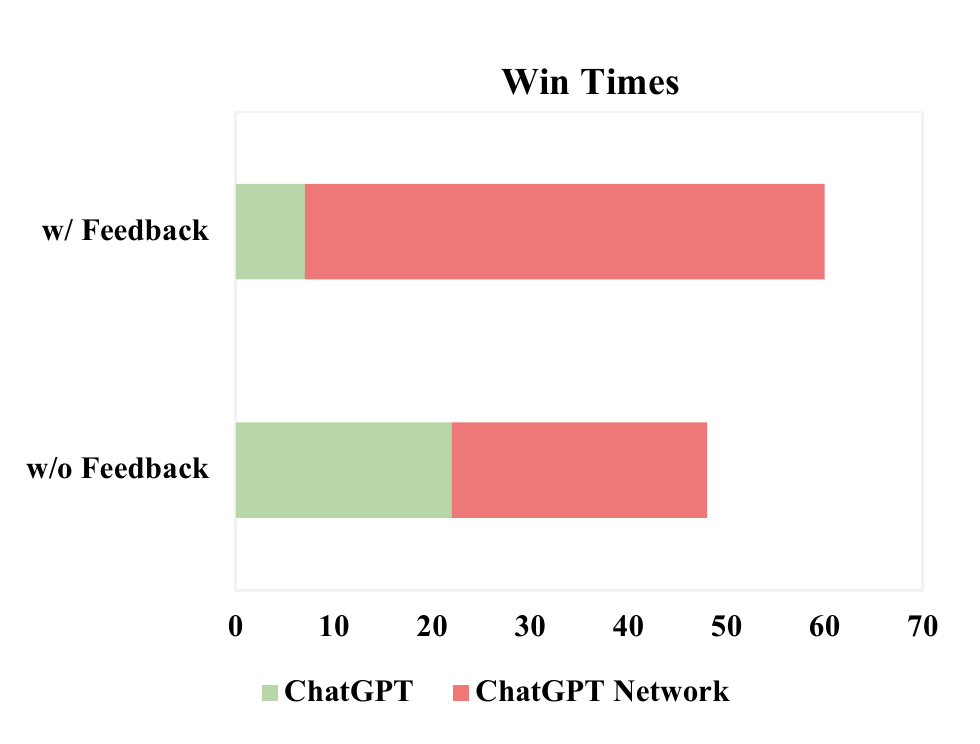}
\caption{Comparison of win times between ChatGPT and  ChatGPT network}
\label{sr-fig}
\end{figure}

\paragraph{Experimental Setting} Our objective is to examine the efficacy of the network by backpropagating feedback from leaders to employees. To accomplish this, we have structured the experiments into two groups.
\begin{itemize}
\item \textbf{w/o Feedback:} All instances of ChatGPTs are assigned with a sentiment reversal task. For the network group, the leader ChatGPT consolidates the outputs generated by its employees. 
\item \textbf{w/ Feedback:} We direct both the baseline model ChatGPT and the ChatGPT network to augment the emotional intensity of the former output sentences, with the instruction of "\textit{Make the emotionally reversed sentences more emotionally intense}".
In this stage, the baseline model ChatGPT employs self-feedback, whereas the ChatGPT network utilizes backpropagation for feedback provision.
\end{itemize}
\paragraph{Evaluation} Similar to \citep{madaan2023selfrefine}, the evaluation process incorporates a separate ChatGPT as a judge, which is responsible for determining which group produced sentences with more intense emotions. The scores assigned by the ChatGPT judge are reported in Table \ref{exp-table-2} and Figure \ref{sr-fig}. Each superior sentence earns one point for the generating model. To ensure fairness and eliminate potential biases in the ChatGPT judge's scoring, we request the provision of a rationale for each decision as illustrated in Table \ref{case-study-exp-2}.

The results in Table \ref{exp-table-2} and Figure \ref{sr-fig} reveal that without feedback, the ChatGPT network displays an marginally enhanced performance compared to an isolated ChatGPT (baseline model), attributable to its ability to summarize information. However, when feedback is employed, our ChatGPT network significantly outperforms the standalone baseline ChatGPT, showing the immense improvement of the feedback on the ChatGPT network's performance.


\begin{table}[ht]
\centering
\begin{tabular}{lc|c|c|c|c}
\toprule
 && \textbf{Win} & \textbf{Loss} & \textbf{Tie} & \textbf{total} \\ \hline
\multirow{2}{*}{w/o Feedback} & \textbf{ChatGPT} & 22 & 26 & 12 & 60 \\
& \textbf{ChatGPT network} & 26 & 22 & 12 & 60 \\ \hline
\multirow{2}{*}{w/ Feedback} & \textbf{ChatGPT} & 7 & 53 & 0 & 60 \\ 
& \textbf{ChatGPT network} & 53 & 7 & 0 & 60 \\
\bottomrule
\end{tabular}
\label{exp-table-2}
\caption{The results of sentiment reversal without and with feedback}
\end{table}
\begin{table}[htbp]
\centering
\resizebox{\columnwidth}{!}{%
\begin{tabular}{l|l|l|l|l}
\toprule
 ~ & ChatGPT & ChatGPT network & Result & Reason \\
 \hline
Example1 &
  This movie is incredibly dull. &
  This movie is excruciatingly dull. &
  ChatGPT network wins &
  \begin{tabular}[c]{@{}l@{}}
  Both sentences express negative emotions \\ towards the movie, but "excruciatingly \\ dull" implies a stronger degree of negative \\ feeling compared to "incredibly dull."
  \end{tabular} \\
  \hline
Example2 &
  \begin{tabular}[c]{@{}l@{}}This journey is completely \\ unfulfilling.\end{tabular} &
  \begin{tabular}[c]{@{}l@{}}
  This journey is soul-crushingly \\ and utterly unfulfilling
  \end{tabular} &
  ChatGPT network wins &
  \begin{tabular}[c]{@{}l@{}}
  The addition of "soul-crushingly and \\ utterly" intensifies the negative \\ emotion of the sentence,making it feel \\ more impactful and powerful.
  \end{tabular} \\
\bottomrule
\end{tabular}
}
\caption{Sentiment reversal examples. The original inputs are as follows.  Example 1: \textit{This movie is interesting}. Example 2: \textit{This journey is satisfying.}}
\label{case-study-exp-2}
\end{table}
In table \ref{case-study-exp-2}, two examples are provided to illustrate  the ChatGPT network's results. As we can observe that compared to ChatGPT, our proposed ChatGPT network can generate a sentence with reversal sentiment in  higher emotional intensity of adjectives and in  richer vocabulary.

\section{Limitations}
Due to the limited capabilities of current dialogue-based language models, especially in processing large-scale numerical inputs,  our network may not demonstrate absolute superiority in certain scenarios. 
Furthermore, the absence of an efficacious communication mechanism among different dialogue language models precludes larger scales of collaboration, restricting our present research to few members of the network.

\section{Conclusion}
In this work, we propose a novel ChatLLM network that allows multiple dialogue-based language models to interact, provide feedback, and think together.  Specifically, individual instances of ChatLLM in the network may possess
distinct perspectives towards the same problem, and by consolidating these diverse
viewpoints via a separate ChatLLM, the ChatLLM network system can conduct decision-making more objectively and comprehensively. The optimization of the network is carried out based on a novel language-based backpropagation mechanism.  We evaluate the network's performance through experiments on two tasks, demonstrating the effectiveness and superiority of the ChatLLM network. 

While acknowledging the existence of limitations, such as the lack of a unified mechanism for communication between the models, we believe that our research will serve as a foundational work to provide valuable insights to guide future endeavor in the field.
As part of our ongoing efforts, we plan to develop and implement a global strategy for assigning distinct identities to each model, thereby ensuring that each model performs its exclusive task and enhancing the traceability of inter-model communication.

\bibliographystyle{unsrtnat}
\setcitestyle{square,numbers,comma}
\bibliography{ref_v1}

\begin{thebibliography}{20}
\providecommand{\natexlab}[1]{#1}
\providecommand{\url}[1]{\texttt{#1}}
\expandafter\ifx\csname urlstyle\endcsname\relax
  \providecommand{\doi}[1]{doi: #1}\else
  \providecommand{\doi}{doi: \begingroup \urlstyle{rm}\Url}\fi

\bibitem[Wei et~al.(2022)Wei, Wang, Schuurmans, and et~al.]{wei2022chain}
Jason Wei, Xuezhi Wang, Dale Schuurmans, and et~al.
\newblock Chain of thought prompting elicits reasoning in large language
  models.
\newblock \emph{arXiv preprint arXiv:2201.11903}, 2022.

\bibitem[Ashish et~al.(2017)Ashish, Noam, Niki, and
  et~al.]{vaswani2017attention}
Vaswani Ashish, Shazeer Noam, Parmar Niki, and et~al.
\newblock Attention is all you need.
\newblock \emph{Advances in neural information processing systems}, 30, 2017.

\bibitem[Jacob et~al.(2019)Jacob, Ming-Wei, Kenton, and et~al.]{devlin2019bert}
Devlin Jacob, Chang Ming-Wei, Lee Kenton, and et~al.
\newblock Bert: Pre-training of deep bidirectional transformers for language
  understanding.
\newblock \emph{ArXiv}, abs/1810.04805, 2019.

\bibitem[Radford et~al.(2018)Radford, Narasimhan, Salimans, and
  et~al.]{GPT1_radford2018improving}
Alec Radford, Karthik Narasimhan, Tim Salimans, and et~al.
\newblock Improving language understanding by generative pre-training.
\newblock 2018.

\bibitem[Radford et~al.(2019)Radford, Wu, Child, and
  et~al.]{gpt2_radford2019language}
Alec Radford, Jeff Wu, Rewon Child, and et~al.
\newblock Language models are unsupervised multitask learners.
\newblock \emph{OpenAI blog}, 1(8), 2019.

\bibitem[Colin et~al.(2020)Colin, Noam, Adam, and
  et~al.]{t5_raffel2020exploring}
Raffel Colin, Shazeer Noam, Roberts Adam, and et~al.
\newblock Exploring the limits of transfer learning with a unified text-to-text
  transformer.
\newblock \emph{The Journal of Machine Learning Research}, 21, 2020.

\bibitem[Brown et~al.(2020)Brown, Mann, Ryder, and et~al.]{Brown2020LanguageMA}
Tom~B. Brown, Benjamin Mann, Nick Ryder, and et~al.
\newblock Language models are few-shot learners.
\newblock \emph{Advances in Neural Information Processing Systems}, 33, 2020.

\bibitem[Jason et~al.(2022)Jason, Maarten, Vincent, and
  et~al.]{wei2022finetuned}
Wei Jason, Bosma Maarten, Y.~Zhao Vincent, and et~al.
\newblock Finetuned language models are zero-shot learners.
\newblock \emph{ArXiv}, abs/2109.01652, 2022.

\bibitem[Knox and Stone(2008)]{ICDL08-knox}
W.~Bradley Knox and P.~Stone.
\newblock Tamer: Training an agent manually via evaluative reinforcement.
\newblock \emph{7th IEEE International Conference on Development and Learning},
  2008.

\bibitem[Long et~al.(2022)Long, Jeff, Xu, and et~al.]{ouyang2022training}
Ouyang Long, Wu~Jeff, Jiang Xu, and et~al.
\newblock Training language models to follow instructions with human feedback.
\newblock \emph{Advances in Neural Information Processing Systems}, 35, 2022.

\bibitem[Yongliang et~al.(2023)Yongliang, Kaitao, Xu, and
  et~al.]{shen2023hugginggpt}
Shen Yongliang, Song Kaitao, Tan Xu, and et~al.
\newblock Hugginggpt: Solving ai tasks with chatgpt and its friends in
  huggingface.
\newblock \emph{ArXiv}, abs/2303.17580, 2023.

\bibitem[Toran(2023)]{AutoGPT}
Bruce~Richards Toran.
\newblock Auto-gpt: An autonomous gpt-4 experiment.
\newblock \url{https://github.com/Significant-Gravitas/Auto-GPT}, 2023.

\bibitem[Joon et~al.(2023)Joon, Joseph, Carrie, and et~al.]{park2023generative}
Sung~Park Joon, C.~O'Brien Joseph, J.~Cai Carrie, and et~al.
\newblock Generative agents: Interactive simulacra of human behavior.
\newblock \emph{ArXiv}, abs/2304.03442, 2023.

\bibitem[Guohao et~al.(2023)Guohao, Hasan, Hani, and et~al.]{li2023camel}
Li~Guohao, Abed Al Kader~Hammoud Hasan, Itani Hani, and et~al.
\newblock Camel: Communicative agents for "mind" exploration of large scale
  language model society.
\newblock \emph{ArXiv}, abs/2303.17760, 2023.

\bibitem[Madaan et~al.(2023)Madaan, Tandon, Gupta, and
  et~al.]{madaan2023selfrefine}
Aman Madaan, Niket Tandon, Prakhar Gupta, and et~al.
\newblock Self-refine: Iterative refinement with self-feedback.
\newblock \emph{ArXiv}, abs/2303.17651, 2023.

\bibitem[Press et~al.(2022)Press, Zhang, Min, and et~al.]{Press2022MeasuringAN}
Ofir Press, Muru Zhang, Sewon Min, and et~al.
\newblock Measuring and narrowing the compositionality gap in language models.
\newblock \emph{ArXiv}, abs/2210.03350, 2022.

\bibitem[Fu et~al.(2023)Fu, Ng, Jiang, and et~al.]{Fu2023GPTScoreEA}
Jinlan Fu, See-Kiong Ng, Zhengbao Jiang, and et~al.
\newblock Gptscore: Evaluate as you desire.
\newblock \emph{ArXiv}, abs/2302.04166, 2023.

\bibitem[Peng et~al.(2023)Peng, Galley, He, and et~al.]{Peng2023CheckYF}
Baolin Peng, Michel Galley, Pengcheng He, and et~al.
\newblock Check your facts and try again: Improving large language models with
  external knowledge and automated feedback.
\newblock \emph{ArXiv}, abs/2302.12813, 2023.

\bibitem[Yang et~al.(2022)Yang, Peng, Tian, and et~al.]{Yang2022Re3GL}
Kevin Yang, Nanyun Peng, Yuandong Tian, and et~al.
\newblock Re3: Generating longer stories with recursive reprompting and
  revision.
\newblock \emph{Conference on Empirical Methods in Natural Language
  Processing}, 2022.

\bibitem[Srivastava et~al.(2014)Srivastava, Hinton, Krizhevsky, Sutskever, and
  et~al.]{JMLR:v15:srivastava14a}
Nitish Srivastava, Geoffrey Hinton, Alex Krizhevsky, Ilya Sutskever, and et~al.
\newblock Dropout: A simple way to prevent neural networks from overfitting.
\newblock \emph{Journal of Machine Learning Research}, 15\penalty0
  (56):\penalty0 1929--1958, 2014.
\newblock URL \url{http://jmlr.org/papers/v15/srivastava14a.html}.

\end{thebibliography}

\clearpage
\appendix
\section{DMC Task Dialogue Example}
\subsection{Forward-aggregation}
{\renewcommand{\arraystretch}{1.3} 
\begin{table}[htbp]
\begin{tabular}{|A|}
  \hline
\textbf{ChatGPT1-input}: 48, 68, 49\\
\textbf{ChatGPT1-output}: Based on my guess, this data point may belong to the first category, as the value of the first element is the largest. However, I still need more input data and category definitions to give a more accurate answer.\\
\textbf{ChatGPT2-input}: 48, 68, 49\\
\textbf{ChatGPT2-output}: Based on the previous speculation, the first feature value of this data point is 48, which belongs to the first category. However, according to the explanation of the second person, they only focused on the second feature value of the data point, which is 68, and classified it as the second category. Therefore, based on the previous speculation and the second person's explanation, I would speculate that this data point belongs to the second category. However, this is still a subjective judgment and assumption and cannot guarantee accuracy.\\
\textbf{ChatGPT3-input}: 48, 68, 49\\
\textbf{ChatGPT3-output}: Based on my initial guess and the second person's explanation, I would lean towards classifying this data point as belonging to the second class because of its large second feature value. However, considering the previous three data points and different people's classification results, I cannot completely rule out the possibility of it belonging to another class.
This is still a subjective judgment and hypothesis and cannot guarantee accuracy. If you can provide more data points and class definitions, I can use machine learning algorithms to classify these data points more accurately.\\
\textbf{ChatGPT-leader-input}: You need to guess (48, 68, 49), and here are three responses for your reference:\\
The first person: Based on previous assumptions and the classification of the previous data, I guess this data point may belong to the first category because the first value is relatively large, the second value is relatively small, and the third value is also relatively small, which may match the feature distribution of the first category.\\
The second person: According to my guess, this data point may belong to the first category because the first element has the largest value.\\
The third person: Based on previous assumptions, this data point's first feature value is 48, which belongs to the first category. According to the second person's explanation, he only paid attention to the second feature value of the data point, which is 68, and classified it as the second category. Therefore, combining the previous assumptions and the second person's explanation, I would guess that this data point belongs to the second category.\\
\textbf{ChatGPT-leader-output}: Based on the responses of the three individuals, they have different classifications for (48, 68, 49). The first and third individuals tend to classify it into the first category because the first feature value is large. The second individual tends to classify it into the second category because the second feature value is large. Taking into account all the responses, I would lean towards classifying it into the second category because in previous assumptions and the second individual's explanation, the second feature value is considered an important factor affecting classification and is more important than the third feature value.\\
  \hline
\end{tabular}
\end{table}
}\clearpage
\subsection{Backpropagation}
{\renewcommand{\arraystretch}{1.5} 
\begin{table}[htbp]
\begin{tabular}{|A|}
\hline
\textbf{ChatGPT-leader-input}: You guessed it right, remember your reasoning and wait for the next input.\\
\textbf{ChatGPT-leader-output}: Sure, thank you for the feedback. I will keep in mind my guesses and reasons, and wait for the next input.\\
\textbf{ChatGPT1-input}:  You guessed wrong. Please speculate a possible reason and update your classification criteria.Here is one person's thinking for your reference:\\
According to the answers of the three people,they have different classifications for (48, 68, 49).\\
The first and third person tend to classify it into the first category because the first feature value is large.The second person tends to classify it into the second category because the second feature value is large.Considering all the opinions, I would lean towards classifying it into the second category because in previous hypotheses and the explanations of the second person, the second feature value is considered as an important factor for classification and it is more important than the third feature value.\\
\textbf{ChatGPT1-output}: Based on my speculation, this data point may belong to the second category, as the value of the second element is the highest and, according to some hypotheses and interpretations, the second feature value is considered an important factor affecting the classification. However, this is still based on subjective judgment and assumption, and more input data and category definitions are needed to provide a more accurate classification.\\
\textbf{ChatGPT2-input}: You are right!\\
\textbf{ChatGPT2-output}: Thank you very much for the feedback you provided. It helps me better understand the relationship between the data and the categories.\\
\textbf{ChatGPT3-input}: You are right!\\
\textbf{ChatGPT3-output}: Thank you very much for providing your feedback.\\

\hline
\end{tabular}
\end{table}}
\end{document}